%% file: main.tex
\pdfoutput=1

\documentclass[11pt]{article}

\usepackage[]{acl}

\usepackage{times}
\usepackage{latexsym}
\usepackage{multirow}
\usepackage{tabulary}
\usepackage{pifont}
\usepackage[inline]{enumitem}
\usepackage[T1]{fontenc}

\usepackage[utf8]{inputenc}

\usepackage{microtype}

\usepackage{todonotes}
\usepackage{comment}
\title{DaLC: Domain Adaptation Learning Curve Prediction for Neural Machine Translation
}

\author{Cheonbok Park$^{\dagger}$, Hantae Kim$^{\dagger}$, Ioan Calapodescu$^{\ddagger}$\\
    \textbf{Hyunchang Cho$^{\dagger}$, and Vassilina Nikoulina$^{\ddagger,*}$}\\
  $^{\dagger}$Papago, NAVER Corp.,$^{\ddagger}$NAVER LABS Europe\\
  \texttt{\{cbok.park,hantae.kim,hyunchang.cho\}@navercorp.com} \\
  \texttt{\{ioan.calapodescu, vassilina.nikoulina\}@naverlabs.com} }

\begin{document}
\maketitle
\begin{abstract}
Domain Adaptation (DA) of Neural Machine Translation (NMT) model often relies on a pre-trained general NMT model which is adapted to the new domain on a sample of in-domain parallel data. Without parallel data, there is no way to estimate the potential benefit of DA, nor the amount of parallel samples it would require. It is however a desirable functionality that could help MT practitioners to make an informed decision before investing resources in dataset creation. We propose a Domain adaptation Learning Curve prediction (DaLC) model that predicts prospective DA performance based on in-domain \textit{monolingual} samples in the source language. Our model relies on the NMT encoder representations combined with various instance and corpus-level features. We demonstrate that instance-level is better able to distinguish between different domains compared to  corpus-level frameworks proposed in previous studies \cite{xia-etal-2020-predicting, kolachina-etal-2012-prediction}. Finally, we perform in-depth analyses of the results highlighting the limitations of our approach, and provide  directions for future research. 
\let\thefootnote\relax\footnotetext{$*$ Corresponding author}
\end{abstract}

\section{Introduction}
\label{sec:introduction}
The classical Domain Adaptation scenario~\cite{freitag2016fast,luongstanford} usually relies on an existing NMT model trained on large datasets originated from various sources. This \textit{general} model is adapted to the new domain, with a small sample of \textit{in-domain parallel} data, through finetuning or other DA methods. Without any parallel in-domain data, we cannot estimate the quality of \textit{general} NMT model on the domain of interest, nor can we  anticipate what will be the benefits of the DA and how much parallel in-domain data is required. 


In this work, we address the problem that translation service providers may regularly face when receiving a request for a new domain translation. In such case, a new domain is often defined by its source language samples, and the translation provider needs to invest into in-domain parallel dataset creation in order to be able to perform evaluation and Domain Adaptation of its \textit{general} model. Current state of the art research in NMT Domain Adaptation rarely provides any insights on the amount of data required to perform Domain Adaptation depending on input domain characteristics. This is however a desirable feature that would allow to 
 \begin{enumerate*}[label=(\roman*)]
     \item estimate the data creation cost (time and money-wise) for the client requesting an adaptation to completely new domain;
     \item make an informed decision on how to distribute fixed data creation budget when there is a need to handle multiple DA simultaneously (as we demonstrate further sharing this budget equally across all domains may not be optimal). 
 \end{enumerate*}
 The goal of this work is to gain better insights on Domain Adaptation dynamics, and provide practical guidelines for such a real-life scenario.

Several studies address the problem of learning curve estimation models~\cite{xia-etal-2020-predicting,ye2021finegrained,kolachina-etal-2012-prediction} of MT or NLP models without actually training those. This is done by training a prediction model which takes corpus-level representation $X$ as an input and predicts the score $y$ for this corpus as an output. In the case of MT learning curves prediction, $X$ would correspond to the parallel data sample (used to train MT model), and $y$ is the BLEU score achieved by an MT model trained on $X$. This means that each training point creation requires training a new MT model, which may become very costly if we want to create a training set of reasonable size. 
In this work, we propose a novel framework to perform the learning of the prediction model at \textit{instance-level}. It can significantly decrease the cost of training samples creation, by leveraging instance-level representations.  


There are multiple factors that would impact DA learning curve: the complexity of the new domain, characteristics of the in-domain samples, baseline performance on the new domain, DA algorithm and its hyperparameters. In this work, we concentrate on the factors related to domain complexity and in-domain sample characteristics. We are particularly interested to understand how much can we get from source only in-domain sample, which corresponds to the real-life scenario. 

 \citet{aharoni-goldberg-2020-unsupervised}  points out that the notion of domain can be fuzzy. They suggest that pretrained language models (LMs) representation could contain rich information about different domains, and have demonstrated that there might be significant overlap between different domains at instance-level.  In the context of NMT Domain Adaptation, NMT encoder representations are better suited to characterize different domains, and to evaluate the difficulty of those domains for the \textit{general} NMT model \cite{del2021translation}. 

The main contributions of this work are:
\begin{enumerate}[label=(\roman*)]
    \item We formulate the problem of DA learning curves prediction as instance-level framework and  demonstrate that instance-level representation favours fine-grained knowledge transfer across different domains thus significantly decreasing the cost of training samples creation;
    \item We propose a prediction model that relies on NMT encoder representations combined with a number of other instance and corpus-level characteristics, computed from the monolingual (source) in-domain sample only;   
    \item  We analyse how far we can go in DA performance prediction based on the source side information only, and outline some limitations of this constraint in Section \ref{sec:analysis}.
\end{enumerate}

\section{Related work}
\label{sec:related_work}
The problem of Learning Curve prediction can be related to a number of different existing problems in Natural Language Processing (NLP) and Machine Learning. In this section, we briefly review few works that have tried to predict Learning Curves (also known as \textit{Scaling Laws}) or domain shift for different NLP tasks. We also overview some works on Active Learning for NMT since those features can also be relevant to our task.

\paragraph{Learning Curve and domain shift prediction.}
There is a number of works that have attempted to predict model's performance without actually executing (and even training) the model in different contexts. 
\citet{elsahar-galle-2019-annotate} predicts the classifier's performance drop under the domain shift problem; \citet{elloumi2018asr} estimates the ASR performance for unseen broadcasts. 
\citet{xia-etal-2020-predicting,ye2021finegrained} study the problem of predicting a new NLP task performance based on the collection of previous observed tasks of different nature. Their proposed models are evaluated on MT among other tasks. 

 Closest to our work,  \citet{kolachina-etal-2012-prediction}  predicts the learning curves for the Statistical Machine Translation(SMT) task. They formulate the task as a parametric function fitting problem, and infer the learning curve relying on set of features based on in-domain source and/or target sample (assuming no in-domain parallel sample is available). 
Most of the above-mentioned works rely on corpus-level score predictions and therefore require large amount of trained MT model instances to generate sufficient amount of training points for the predictor model.

There is very recent interest around modeling the \textit{Scaling Laws} for a language model \cite{kaplan2020scaling} or a NMT model \cite{gordon-etal-2021-data, ghorbani2021scaling}. Similar to \cite{kolachina-etal-2012-prediction} these works try to derive a parametric function that would allow to make a connection between the model's training characteristics (amount of data, parameters or compute) and model's final performance.  These models operate at corpus level and do not address the problem of Domain Adaptation (and the fact that different domains may follow different scaling laws functions). 

To best of our knowledge, our work is the first attempt to specifically solve learning curves prediction in the context of NMT Domain Adaptation. We propose an instance-level framework relying on NMT encoder representations (which none of the previous work did) in combination with other features. Owing to this framework, our proposed framework requires a small amount of trained MT models with less than 10 models.

\paragraph{Active learning for NMT.}
Active Learning (AL) algorithms are built to select the most useful samples for improving the performance of a given model. Therefore, the criteria used by AL algorithms in NLP or MT tasks \cite{zhang2017active, zhao-etal-2020-active,peris-casacuberta-2018-active, dou-etal-2020-dynamic, wang2020learning,Zhan_Liu_Wong_Chao_2021} could also serve as discriminative features when it comes to predicting the future NMT performance. We reuse some of the scoring functions introduced by previous work as specified in Section \ref{sec:features}. On the other hand, we believe that a successful Learning Curves prediction framework can help identify important features and/or samples for an AL framework. 





\section{Analysis of real Domain Adaptation Learning curves}
\label{sec:data}
In this section, we analyse several real learning curves for NMT Domain Adaptation to show how learning curves do behave across domains and to motivate our work. 
In order to perform this analysis, we first present the NMT baseline model (i.e., \textit{general} model) we utilize, as well as datasets used for Domain Adaptation. We also discuss the NMT evaluation metrics that we can rely on when training the learning curve predictor model\footnote{In what follows we may refer to the \textit{learning curve predictor model} as \textit{predictor} to avoid the confusion with NMT model.}. 
    
\paragraph{NMT model.}
We consider two different NMT systems: English-German and German-English systems trained on WMT20 dataset~\cite{barrault-etal-2020-findings}. We provide technical details about architecture and datasets used to train those NMT systems in the Appendix \ref{app:mt}. 

\paragraph{Domain Adaptation data.}
 
  We rely on the dataset released by \citet{aharoni-goldberg-2020-unsupervised}. The dataset consists of train/dev/test with deduplicated sentences. This dataset splits for 5 domains (Koran, IT, Medical, Law, Subtitles) from OPUS~\cite{tiedemann-2012-parallel} for German-English. For each domain, we create random samples $S_{n,d}$ of size $n$  (with $n=$1K, 10K, 20K, 100K) at $d$ domain. Those samples are then used to train instances of Domain Adapted models $M_{S_{n,d}}$ resulting in total 19 models\footnote{We didn't train 100K-sampled DA model for \textit{Koran }domain since it only has 20K parallel sentences in the training split}. We will further refer to  the size of these different samples as \textit{anchor points} (of the learning curve). 
  \paragraph{Domain Adaptation.}
For each \textit{general} NMT model (\textit{baseline}), we create a set of Domain Adapted models trained on different samples of in-domain data described previously (anchor points). Domain Adaptation is done via finetuning only with an in-domain dataset (Appendix \ref{app:mt} provides details).


  \paragraph{NMT evaluation.} Our main goal is to obtain a corpus-level score that would allow us to assess the DA performance which is traditionally measured by BLEU score \cite{papineni-etal-2002-bleu}.  While BLEU score may exhibit  reasonable correlation with human judgements at corpus-level, it is known to have poor correlation at instance-level. 
  Recall that we are interested to exploit instance-level representations to favour knowledge transfer across domains, therefore we require a reliable instance-level metric to create gold annotations that the predictor model could learn from. We rely on \textit{chrF} \cite{popovic-2015-chrf} score that,  according to WMT 20 MT evaluation track \cite{mathur-EtAl:2020:WMT}, provides reasonable correlation with human judgement for instance-level evaluations.
  For learning curves prediction we rely on \textit{mean chrF} (average instance-level chrF across the whole test set) as a proxy for corpus-level score thus making connection with instance-level scores used for training of the predictor. 
  
  
\begin{figure}
    \centering
    \includegraphics[width=\columnwidth]{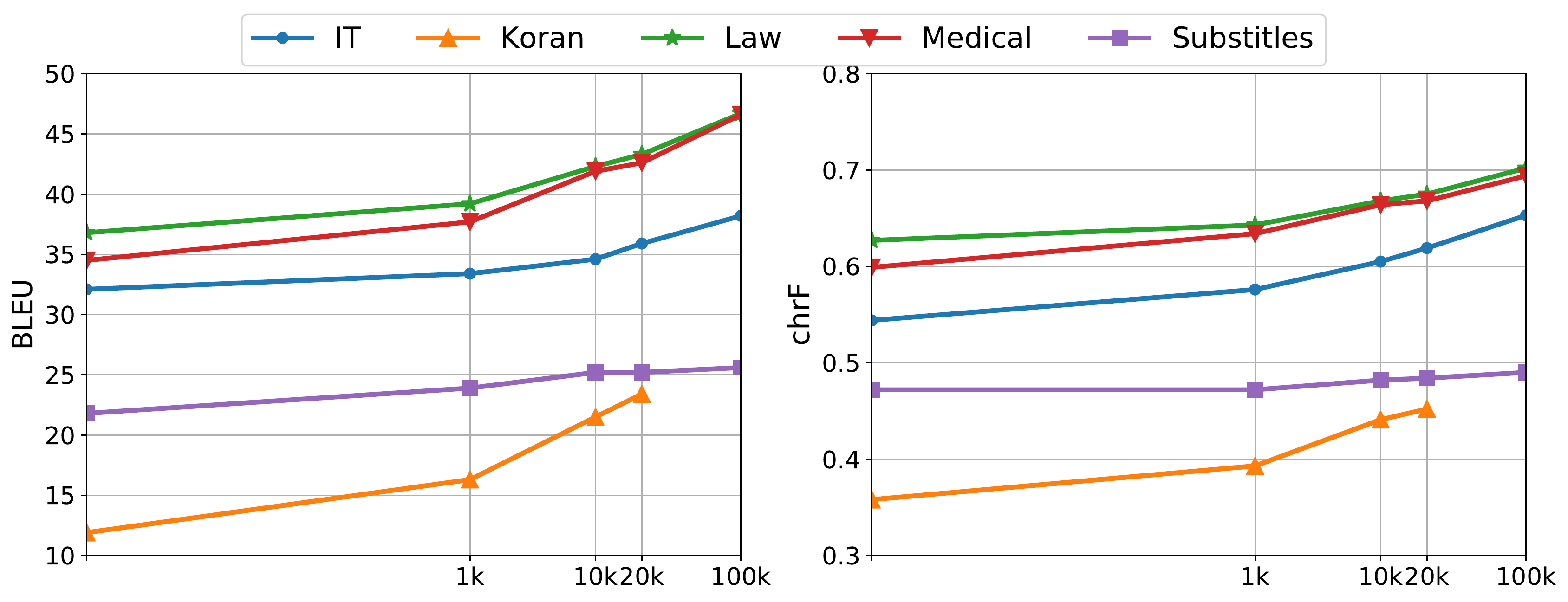}
    \caption{Domain Adaptation Learning Curves for English-German NMT model.  Y-axis represents the performance of Domain adapted model (e.g., chrF or BLEU), X-axis: amount of in-domain samples used by Domain Adaptation (at log scale). }
    \label{fig:gold_curves}
\end{figure}

\paragraph{Analysis of Learning Curves.}

Figure \ref{fig:gold_curves} reports how BLEU and \textit{mean chrF} scores progress with amount of in-domain samples used for Domain Adaptation (English-German). First, we note that \textit{mean chrF} metric exhibits the same behaviour as BLEU when tracing the Learning Curve\footnote{ Table \ref{tab:da_bleu_chrf} in Appendix \ref{app:mt} provides BLEU/chrF for German-English and confirms the above observation. All the evaluations are performed with SacreBLEU toolkit \cite{post-2018-call}.}.


Second, we note the difference in a Domain Adaptation progress for each of domains. For instance, \textit{Koran} learning curve is quite steep, while \textit{Subtitles} domain improves very slightly and reaches a plateau performance at already 1K Domain Adaptation anchor point. We also note that this behaviour isn't necessarily linked to the performance of the \textit{baseline} model on those domains: \textit{Law} and \textit{Medical} domains achieve quite high scores when translated with the \textit{baseline} NMT model (tables \ref{tab:da_bleu_chrf}, \ref{tab:da_stats} in the appendix), but they keep improving as in-domain data grows.  \textit{Subtitles} on the other hand has lower \textit{baseline} BLEU and \textit{mean chrF} scores and Domain Adaptation with growing in-domain data doesn't seem to help much. 
Such behaviour probably reveals the limitations of the dataset, and/or a Domain Adaptation method used for this domain that would merit further investigations.  The problem that we want to address in our work is whether it is possible to forecast such behaviour in advance, and how far we can go in this task with source only in-domain data sample. 

\section{Approach}
\label{sec:approach}
In this section, we first formalize the DA learning curve predictor. Then, we describe the representations and our model for the problem. 




\subsection{Problem setup}
We are given a \textit{baseline model} $M_{G}$ (trained on general corpus $G$), input sentence $x$ and a new domain $d$ defined by its sample $S_d$. The \textit{DA learning curve predictor} can be modeled as a scoring function  $g_{\theta}$ which depends on the instance-level representation $\phi(x)$, and corpus-level representation $\xi(S_d^s)$\footnote{$S_d^s$ denotes source side of in-domain sample $S_d$ since we restrict ourselves to the case where we can only access source side of in-domain samples.}. 

 
 Model $M_{S_d}$ is an NMT model obtained by adapting $M_{G}$ to the new domain(i.e., $d$ on the in-domain sample of parallel sentences $S_d$).  
 Learning of \textit{DA learning curve predictor} can be done by regressing the actual scoring function $y=s(x, M_{S_d})$ that provides translation quality score for an input sentence (from the test domain) $x$ translated with $M_{S_d}$.  $s(x, M_{S_d})$ refers to \textit{chrF} score as discussed in Section \ref{sec:data}.

 The  learning objective is then formulated as    
    

$$    \min_{\theta} \sum_{d \in D} \sum_{x \in T_d} (s(x, M_{S_d}) - g_{\theta}(\phi(x),\xi(S_d^s)))^2$$ 

where $D$ and $T_{d}$ are a set of training domains\footnote{We consider the scenario where the test domain is not known during the predictor learning phrase and disjoint form training domains.} and training sentences of $d$ domain\footnote{ $T_{d}$ does not contain the training sentence for adapted model.}, respectively.


\begin{figure}[]
    \centering
    \includegraphics[width=0.9\columnwidth]{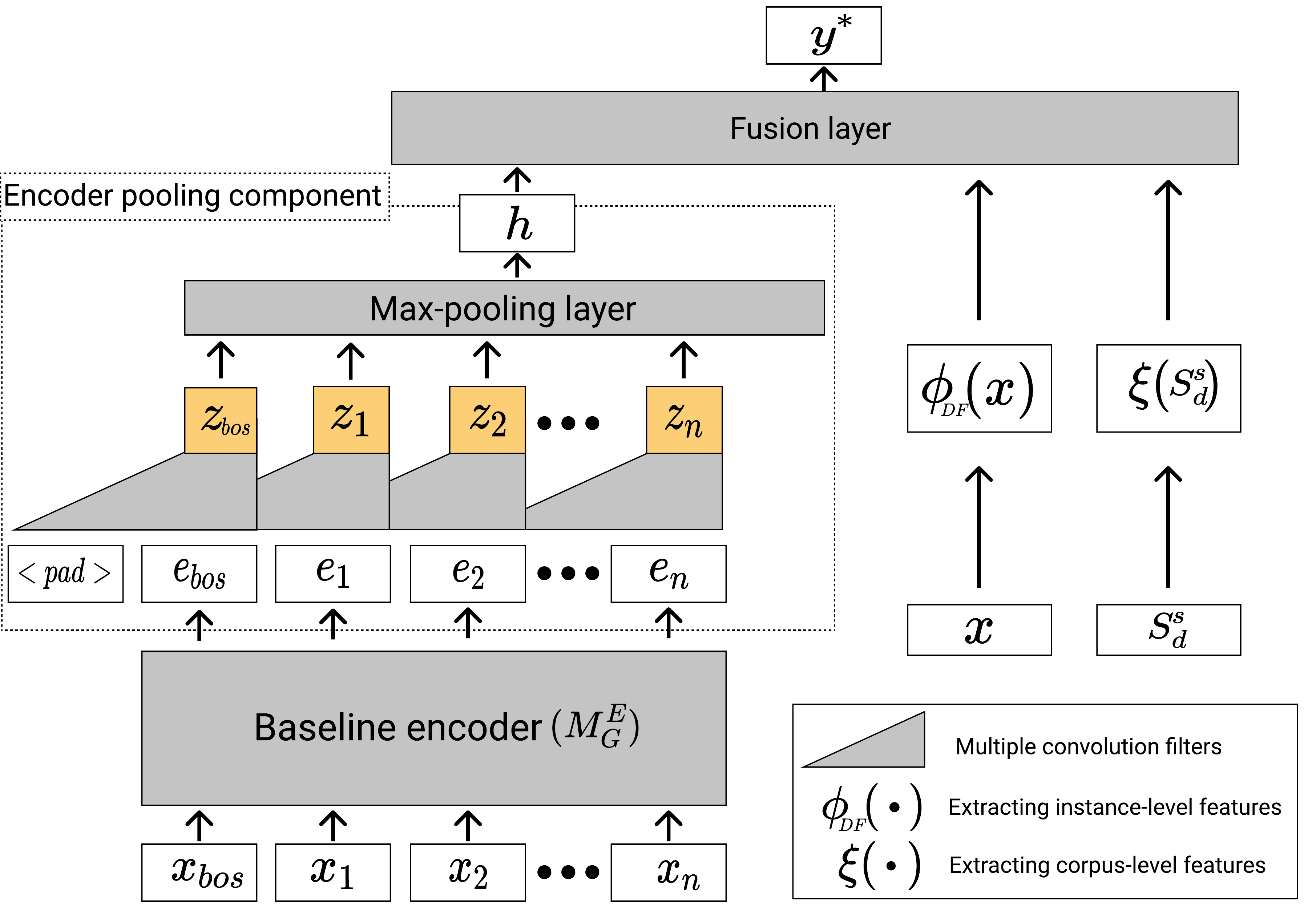}
    \caption{Overview of the learning curve predictor model, where $x_{i}$ is the $i$-th token of $x$. $M_{G}^{E}$ denotes the encoder of the baseline translation model($M_{G}$)}
    \label{fig:arch}
\end{figure}

\begin{figure*}[ht!]
\centering
    \includegraphics[scale=0.32]{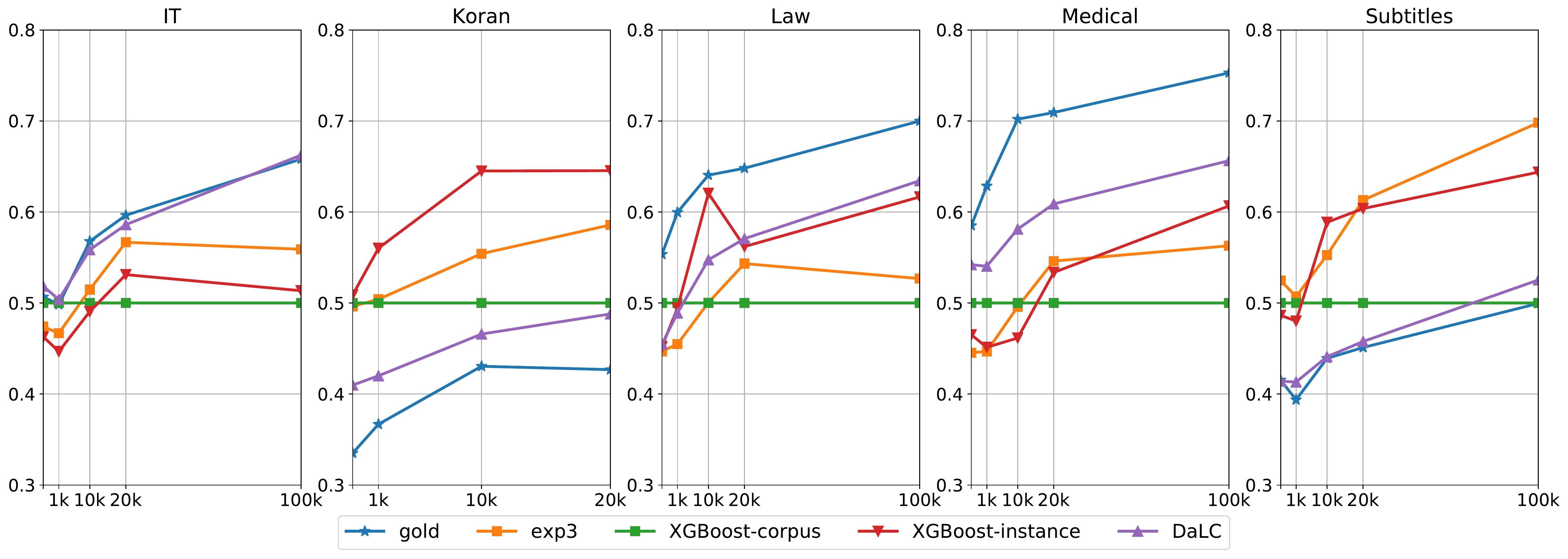}
    \caption{Learning curves provided by DaLC compared against \textit{gold} predictions,  baseline predictors ($exp_3$, XGboost-corpus, XGboost-instance) for German-English/FT.}
    \label{fig:lc}
\end{figure*}

\subsection{Input representations}
\label{sec:features}
In this section, we describe different features we consider for input sentence $x$ and for the source-side of in-domain sample  $S_d^{s}$. 
\paragraph{NMT encoder representations} contain a wealth of information that can be very relevant to the \textit{adaptability} of the model to the new domain. Therefore, we consider it as an important building block of our predictor model. In our implementation, we take the last encoder layer representation of each token, which are then aggregated through a pooling component in a single vector $\phi_{enc}(x)$ (\textit{Encoder pooling component} at Figure \ref{fig:arch}).
 
\paragraph{Corpus-level features} $\xi(S_{d}^{s})$ allow us to characterize the in-domain sample $S_{d}^{s}$ with respect to its size, diversity, and similarity to pretrained data $G$.  
In the simplest case we consider sample size (amount of instances) as a single corpus-level feature. 
In addition we add some of the features used by  \citet{xia-etal-2020-predicting, kolachina-etal-2012-prediction}  namely: (1) amount of tokens in $S_{d}^{s}$; (2) vocabulary overlap ratio between $G$ and $S_{d}^{s}$; (3) the average sentence length (in characters, in tokens); (4) the number of unique tokens in $S_{d}^{s}$; and (5) type token ratio~\cite{richards1987type}\footnote{We apply log-scale to all feature values to reduce the variability in feature values.}. 
 
 
\paragraph{Instance difficulty features (DF).} The quality of the translation depends on how difficult input sentence is for the NMT system.

 Difficulty features $\phi_{DF}(x)$ include model-based uncertainty functions from \citet{zhao-etal-2020-active}: (1) least-confidence score, (2) margin score (3) average token entropy. Those features rely on the pretrained model $M_G$, and therefore could be seen as redundant with encoder representations. However, as we demonstrate in Section \ref{sec:results} they  turn out to be helpful in some cases. Finally, we consider (4) the cross-lingual cosine similarity score between representation of the source sentence $x$ and its translation $M_G(x)$. These representations are obtained from an external pretrained multilingual sentence embedding model, LaBSE~\cite{feng2020language}.

\subsection{Domain Adaptation Learning Curve (DaLC) predictor}
DaLC predictor corresponds to the model depicted by Figure \ref{fig:arch}. 
It contains two main components: 
\begin{enumerate*}[label=(\arabic*)]
    \item \textit{encoder pooling component} that processes NMT encoder representations (given as a sequence of vectors) and produces a single vector $\phi_{enc}(x)$. 
    \item \textit{fusion layer} combining encoder representations $\phi_{enc}(x)$ with other pre-computed instance-level features $\phi_{DF}(x)$ and corpus-level features $\xi(S_d)$. 
\end{enumerate*}

 In our experiments, we use the multi-filter CNN architecture proposed by \citet{kim-2014-convolutional} that is widely used in text classification tasks as \textit{encoder pooling component}. For\textit{fusion component}, we simply stack $K$ feed-forward layers followed by ReLU and Sigmoid activation functions at the final layer.



%

\begin{table*}
\small
    \centering
    \begin{tabular}{|c|c|c|c|c|c|c|}
    \hline
         & IT & Koran & Law & Medical & Subtitles & Avg \\\hline
         & \multicolumn{6}{c|}{German-English/FT} \\\hline
        \textit{exp3} & 0.125 & 0.292 & 0.305 & 0.397 & 0.321 & 0.288 \\
        XGboost-corpus & 0.197 & 0.235& 0.305& 0.435& 0.157&  0.266\\\hline
        XGboost-instance & 0.084 & 0.201 & 0.062& 0.176& 0.126& 0.130 \\
        DaLC & \textbf{0.009} & \textbf{0.058} & \textbf{0.057} & 0.094& \textbf{0.015}& \textbf{0.047} \\
        DaLC $/$ DF  &0.011 & 0.065  & 0.058& 0.117 & 0.022& 0.055  \\
        DaLC $/$ corpus  & 0.049& 0.045 & 0.097 & 0.117& 0.052& 0.072 \\ 
        DaLC $/$ NMTEnc  & 0.025& 0.148& 0.085& \textbf{0.081}&  0.061& 0.080\\ 
        \hline\hline
         & \multicolumn{6}{c|}{English-German/FT} \\\hline
         \textit{exp3} & \textbf{0.035} & 0.180& \textbf{0.116}& 0.114& 0.112& 0.112 \\
        XGboost-corpus & 0.111& \textbf{0.081}& 0.169& 0.169& \textbf{0.029}& 0.112 \\\hline
        XGboost-instance & 0.072& 0.157& 0.159& 0.116& 0.09& 0.119 \\
        DaLC & 0.048& 0.107&0.123& 0.041& 0.057& 0.075\\ 
        DaLC $/$ DF  & 0.065& 0.102&0.123& 0.044& 0.053& 0.077\\ 
        DaLC $/$ corpus  & 0.048& 0.086& 0.126& 0.043& 0.063& \textbf{0.073}\\ 
        DaLC $/$ NMTEnc  & 0.043 & 0.169& 0.125& \textbf{0.016}& 0.095& 0.090 \\ 
        \hline\hline
    & \multicolumn{6}{c|}{German-English/Adapter} \\\hline
        \textit{exp3} & 0.055  &  0.175& 0.100 & 0.169 & 0.141 &0.128\\
        XGboost-corpus &  0.079&0.137& 0.115 &0.166 &0.083 & 0.116 \\\hline
        XGboost-instance &0.100& 0.145 & 0.092 &0.162 & 0.097& 0.119 \\
        DaLC &  \textbf{0.022}& \textbf{0.045}& \textbf{ 0.080}& 0.102& \textbf{0.019}& \textbf{0.054}\\
        DaLC $/$ DF  & 0.024& 0.057&  0.082& 0.109& 0.020& 0.058  \\
        DaLC $/$ corpus & 0.030& 0.048& 0.087& 0.109& 0.036& 0.062\\ 
        DaLC $/$ NMTEnc  &0.079& 0.168& 0.081& \textbf{0.068}& 0.089& 0.097\\ 
        \hline
    \end{tabular}
    \caption{RMSE for DaLC predictor compared against \textit{exp3} and XGboost baselines on De-En and En-De  directions, where FT and Adapters correspond to the NMT Domain Adaptation method in each experiment. 
    }
    \label{tab:full_results_ende}
\end{table*}

\section{Experimental settings}
\label{sec:experimentaion_setting}

\subsection{Data and Evaluation}
\paragraph{English-German.}We rely on data described in Section \ref{sec:data}. 
We use development split (2K sentences) for \textit{predictor training}: it is randomly split into train (80\%) and validation (20\%) sets. 
 
 The predictor is evaluated on test split portion for each domain (2K sentences). The results reported for each domain are obtained with the predictor trained in Leave-one-out settings (e.g., predictor trained on\textit{ Law}, \textit{IT}, \textit{Koran}, \textit{Medical} is evaluated on \textit{Subtitles}). Such evaluation allows us to mimic real-life scenario where we need to predict performance for a completely new domain which is not known at training time. 

The evaluation of the predictor is done by measuring error between the predicted score and the ground truth score (measured by \textit{mean chrF}) across all the anchor points. Following \citet{kolachina-etal-2012-prediction}, we report Root Mean Square Error (RMSE) across all the available test anchor points. In addition, we report absolute error at each anchor point (when possible) to allow for finer-grained analysis of the results. Each experiment is repeated 5 times with different random seeds, and an average across all runs is reported.
 
 \paragraph{English-Korean.} We consider  five specialized domains (\textit{technology}, \textit{finance}, \textit{travel}, \textit{sports} and \textit{social science}) publicly available from AI-Hub~\footnote{\url{https://aihub.or.kr/aihub-data/natural-language/about}}. The size of the validation and test sets are 10k and 5k, respectively. We randomly sample  $S_{n,d}$ of 2K sentences following the settings of  English-German experiment. Detailed information of baseline NMT models and DA models for English-Korean is described in Appendix~\ref{app:mt}. We adopt the same evaluation scenario as in English-German (leave-one-out settings with RMSE evaluation)

\subsection{Baseline predictor models}
\label{sec:baselines}
Traditionally predictor models are evaluated against a naive baseline predicting the \textit{mean} over observations used for training.  However, such baseline does not make much sense in the context of learning curve prediction since it is unable to extrapolate to new anchor points.  

\paragraph{\textit{exp3} baseline.}
\label{sec:exp3}
$exp_3$ is a 3-parameter function that is defined by $y=c-e^{(-ax+b)}$. \citet{kolachina-etal-2012-prediction} has identified this function as a good candidate for SMT learning curve prediction fitting. In our experiment, we fit this function through least-squares algorithm to all the observations we dispose across all the domains and anchor points (19 points). This function can be seen as an extension of \textit{mean} baseline allowing to extrapolate to unobserved anchor points.

\paragraph{XGboost-based baselines.}
Following \citet{xia-etal-2020-predicting} we also use gradient boosting trees model \cite{friedman2000greedy}, implemented in XGboost \cite{xgboost}. 
\textit{XGboost-corpus} baseline corresponds to the XGboost model trained with the corpus-level features: this baseline is comparable with the one used by \citet{xia-etal-2020-predicting}. We also compare our model against \textit{XGboost-instance} baseline which corresponds to the XGboost model trained with the full set of features (section \ref{sec:features}) that DaLC predictor is trained with. NMT encoder representations are squeezed in a single vector via min-max pooling and fed to XGboost along with other instance-level features. Comparing \textit{XGBoost-corpus} and \textit{XGboost-instance} results allows us to decouple the impact of instance-level representations and impact of the predictor model learnt from these representations. Exact details of XGboost training are reported in Appendix \ref{app:xgboost}. 
\begin{table*}
    \centering
    \begin{tabular}{|c|c|c|c|c|c|c|}
    \hline
         & Finance & Social & Sports & Tech & Travel & Avg \\\hline
         & \multicolumn{6}{c|}{Korean-English/FT} \\\hline
        \textit{exp3}    &  0.055& 0.028 & 0.018& \textbf{0.080}    & 0.069 & 0.050 \\
        XGboost-corpus   &  0.246&  0.187 & 0.205  & 0.307&  0.152 & 0.219 \\\hline
        XGboost-instance &  0.035& 0.066 & 0.031 &0.092& 0.027  & 0.050 \\
        DaLC             & 0.028 & \textbf{0.007}& \textbf{0.016}   & 0.085 &\textbf{0.010} &\textbf{0.029} \\
        DaLC $/$ DF      &  0.046& 0.010& 0.020 & 0.086&  \textbf{0.010}  & 0.034 \\
        DaLC $/$ corpus  & 0.037& 0.009 & 0.019 & 0.104 & 0.017 & 0.037\\ 
        DaLC $/$ NMTEnc  &  \textbf{0.025} &0.046&  0.021& 0.089&  0.011  & 0.038\\ 
        \hline
    \end{tabular}
    \caption{RMSE for DaLC predictor compared against \textit{exp3} and XGboost baselines on the finetuning method with the Ko-En direction. 
    }
    \label{tab:full_results_enko_FT}
\end{table*}
\subsection{DaLC predictor} 
In our preliminary experiments, we observed that the capacity of \textit{encoder pooling component} does not have much impact on the overall performance.  We believe this is because NMT encoder outputs already provide rich contextualized representations of the input sequence. On the other hand, it is important to give enough capacity to the \textit{Fusion layer} which should mix instance-level representations (including NMT encoder representation) with corpus-level features. In our experiments, \textit{encoder pooling component} is a single layer multi-filter CNN (with 3 filters of size 2, 3, and 4).  \textit{Fusion layer} is composed of 4 feed-forward layers of hidden size 512, followed by ReLU activation, and final feed-forward layer followed by Sigmoid activation that brings the final prediction at 0-1 scale.
We use Mean Squared Loss (MSE) for training. We apply early stopping criteria with the patience of 10 epochs. 
We provide more training hyperparameters in the Appendix \ref{app:implem}.



\section{Results}
\label{sec:results}
\label{result_discussion}

\subsection{Prediction for observed anchor points}
Table \ref{tab:full_results_ende} reports the results across different domains for English-German and German-English.  Table \ref{tab:full_results_enko_FT} reports the results for Korean-Enligsh. We compare corpus-level baselines (\textit{exp3} and \textit{XGboost-corpus}) against different instance-level predictors:
\begin{enumerate*}[label=(\roman*)]
    \item our \textit{DaLC predictor} relying  on the full set of instance- and corpus-level features described in the section \ref{sec:features};
    \item \textit{XGboost-instance} (section \ref{sec:baselines}) relying on same features as DaLC;  
    \item Ablation of different groups of features from full model (DaLC/DF, DaLC/corpus or DaLC/NMT encoder)
\end{enumerate*}
    
For each domain, we report RMSE when comparing predicted \textit{mean chrF} to the gold \textit{mean chrF}\footnote{The gold mean chrF corresponds at anchor point $K$ for domain $d$ corresponds to the actual value of \textit{mean chrF} obtained after adaptation of NMT model with $K$ samples from domain $d$.} score averaged across all the anchor points (0, 1k, 10k, 20k and 100k). 

\begin{figure*}[!th]
    \centering
    \includegraphics[width=\linewidth]{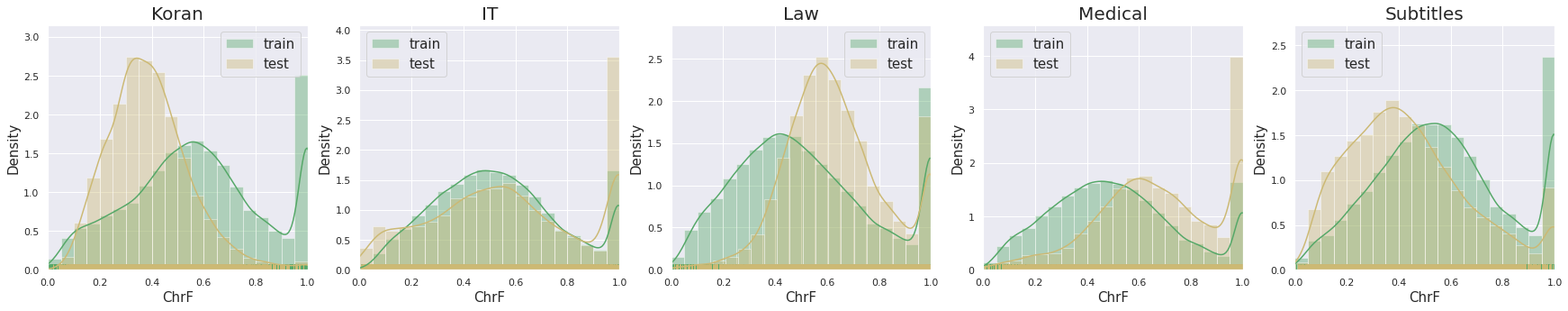}
    \caption{ChrF score distribution of training and test datasets across each experiment case in the German-English FT experiment. Each histogram shows the difference between the training and test ChrF score distributions in each leave one out setting. For instance, the ChrF score distribution of the Koran column obtains where training distribution denotes all training anchor points of IT, Law, Medical and Subtitles domain, and test distribution represent chrF score in all anchor points of the Koran domain. }
    \label{fig:dist_de_en}
\end{figure*}
\paragraph{Instance-level vs corpus-level.}

We note that instance-level models generally outperform corpus-level models for most of domains and language pairs. An exception is English-German direction, where \textit{XGboost-corpus} leads to better prediction then \textit{XGboost-instance} in \textit{Koran} and \textit{Subtitles} domain. According to additional visualisation of these result in  Figure \ref{fig:lc} and Figure \ref{fig:lc_ende} (in Appendix) we see that \textit{XGboost-corpus} model actually fails to learn  meaningful patterns as it predicts the same score (0.5) for all the domains across all the anchor points\footnote{We believe this might be due to the very small amount of corpus-level training samples.}.  It leads to lower RMSE for \textit{Koran} and \textit{Subtitles} domains only because the gold \textit{mean chrF} for those sets is very close to 0.5. Therefore, even if the instance-level leads to higher RMSE it provides more accurate predictions overall as shown by \textit{Avg} column in Table~\ref{tab:full_results_ende}. 

We observe that  while DaLC reaches lowest RMSE across all the domains for German-English and Korean-English, it is not necessarily true for English-German. We note however, that DaLC performance varies less across domains, and reaches overall best performance (reflected by \textit{Avg} column) which means that it is less influenced by overall \textit{mean} performance (as opposed to corpus-level models),  and is able to  better exploit instance-level representations that favour knowledge transfer across domains.

In addition in the Appendix \ref{app:comp_cost} we report computation cost for instance-level and corpus corpus level models.



\paragraph{Impact of different features.} We note that the impact of DF and corpus features varies across the domains. One clear trend is that NMT encoder features seem to be important for predictor quality. An interesting exception is the \textit{Medical} domain, where the removal of NMT encoder representations seem to reach the best RMSE. Furthermore, Appendix \ref{sec:abblation} provides an in-depth ablation study for each feature used in the model.


\paragraph{Impact of DA algorithm.} 
We examine the effectiveness of the proposed method on the different adaptation algorithm, comparing adapter layers~\cite{bapna-firat-2019-simple} to full finetuning.  The Adapter layer is a small module inserted on top of  each encoder and decoder block and updated only with in-domain samples (while keeping the rest of the model frozen). The hidden dimension size of the adapter is 1024. Details of the adapted models is provided in Table~\ref{tab:da_bleu_chrf}. 

Table~\ref{tab:full_results_ende} demonstrates the quality of predictions domain adapted model via Adapter layers in the De-En direction. Comparing to FT results on the De-En direction, we notice that the predictor quality behaves similarly for both DA methods. It indicates that DaLC can be extended to other DA methods.

\paragraph{Impact of other factors.} We report additional experiments such as depending on the number of domains for training predictor and performance on a mixing of two different domains in Appendix \ref{sec:extra_dom} and Appendix \ref{app:mix-ko-en}. 
\subsection{Interpolation and Extrapolation of DA performance}

One of practical and possible scenarios is prediction of DA performance for the anchor points that were not observed in the training data. Table~\ref{tab:extrapolation_res} shows the accuracy of our predictor for interpolation\footnote{Interpolation: prediction for the unseen anchor points that are within the range of observed anchors.} ($3k$ and $40k$) and extrapolation\footnote{Extrapolation: prediction for the anchor point that lies beyond observed anchors.} ($160k$) scenarios for \textit{Subtitles} test domain in the De-En direction. We recall that the predictors have been trained on 0k, 10k, 20k and 100k anchor points for \textit{IT}, \textit{Medical}, \textit{Law} and \textit{Koran} domains. 

We report the absolute error with respect to gold \textit{mean chrF} (MAE) for these specific anchor points. We can see that DaLC achieves significantly lower error compared to other baselines. We note however that extrapolation error ($160k$) is higher compared to interpolation errors ($3k$ and $40k$).

\begin{figure*}[!t]
    \centering
    \includegraphics[width=\linewidth]{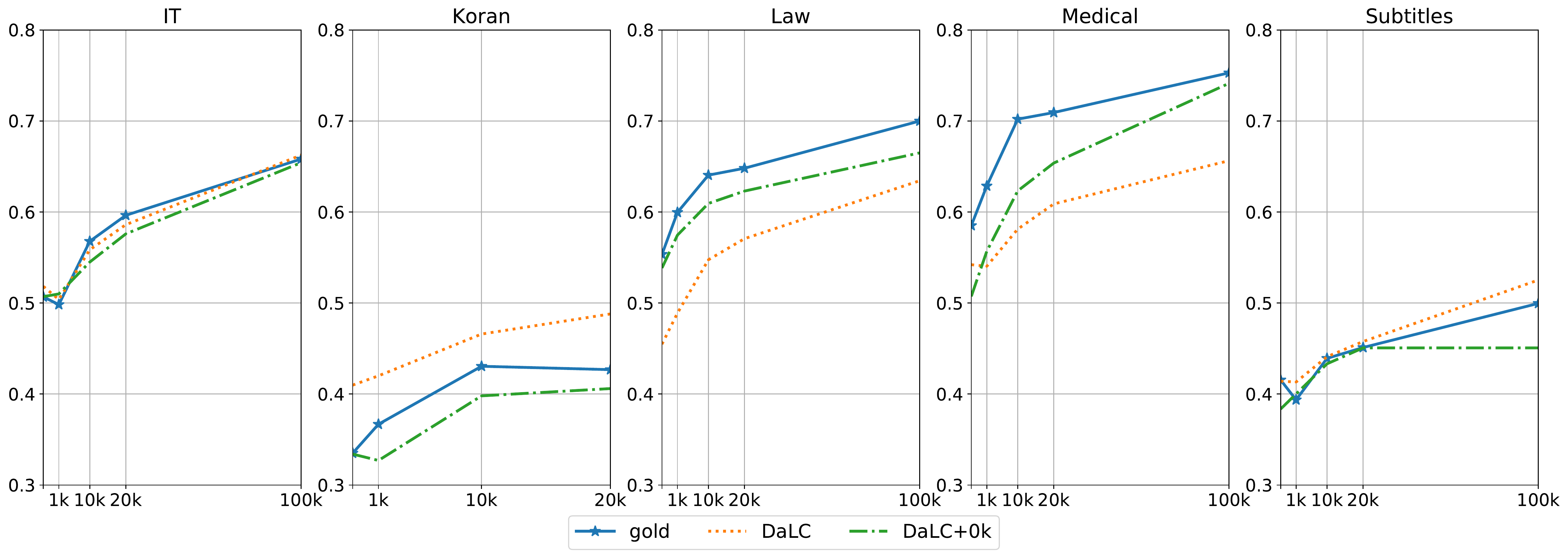}
    \caption{Comparison of learning curve prediction with DaLC and DaLC \texttt{+with 0k} on the De-En direction.}
    \label{fig:dalc_0k}
    \vspace{-0.06in}
\end{figure*}    
\begin{table}[]
\small
    \centering
    \begin{tabular}{c|c|c|c}
\hline
model &  3K & 40k & 160k \\\hline
    $exp_3$ & 0.1087 & 0.1640 & 0.1934 \\
    XGboost-instance & 0.1546 &0.0922 & 0.1125\\
         DaLC &\textbf{0.0063} & \textbf{0.0080}& \textbf{0.0413} \\
         \hline
         
    \end{tabular}
    \caption{Results for interpolation and  extrapolation of learnt models to new  (unseen in training) anchor point: we report the absolute value of the difference between gold and predicted values (subtitles domain)}
    \label{tab:extrapolation_res}
\end{table}

\section{Analysis and Discussion}
\label{sec:analysis}

The results reported in previous section suggest that overall DaLC prediction error  depends a lot on the nature of the test domain (as seen on Figures \ref{fig:lc}, \ref{fig:lc_ende} and Table \ref{tab:full_results_ende}). Thus,  all the predictors tend to overestimate the \textit{mean chrF} score on \textit{Koran} domain, or underestimate the score on \textit{Law} domain. In this section we try to analyse this phenomenon and explain such behaviour.

\subsection{Train/test data distribution}
\label{sec:distribution}
Figure \ref{fig:dist_de_en} provides visualization of training and test distribution for each domain. 
We can see that in the case of \textit{Koran} and \textit{Law} domains there is a highest shift between train and test distribution of chrF scores: \textit{Koran} has more low-quality translations (low chrF value) compared to its training domains, while for \textit{Law} it is the opposite. This discrepancy leads to underestimated scores for \textit{Law} domain and overestimated scores for \textit{Koran}. On the other hand, we can see the predictor has higher accuracy in other domains that have a similar distribution between training and test distributions. We observe the same patterns for the En-De direction (Figure \ref{fig:dist_en_de}), or when extending our framework to more training domains (Appendix \ref{sec:extra_dom}). 

 


\subsection{Adding 0 anchor point}
\label{sect:with_0k} The above finding regarding train/test discrepancy leading to a higher prediction error implies that even though our model is able to exploit instance-level representations to some extent (it achieves lower prediction error compared to corpus-level baseline), it is still heavily impacted by overall data distribution.  One possible explanation for such behaviour would be the fact that source-level representations may not contain enough information for model to rely on to predict future translation quality. 
A simple example would be a source sentence that should be either translated in ``formal'' or ``informal'' manner depending on what the target domain is. When the model lacks information about target language distribution for the new domain its simply has no mean to learn faithful predictor. 
  
 In this experiment, we consider the scenario where we have access to a small sample of parallel data (2K sentences) for the test domain. When such sample is available the predictor training data can be enriched with 0-anchor point samples corresponding to the translations produced by our baseline model and their corresponding chrF scores. 

Figure \ref{fig:dalc_0k} and Figure \ref{fig:dalc_0k_ende} in Appendix demonstrate the results of this experiment (Table \ref{tab:with_0k_anchor} reports RMSE scores for the reader interested in more in-depth analysis). We see that addition of 0k anchor point significantly improves the learning curves obtained by DaLC predictor.  This confirms our hypothesis that relying on monolingual in-domain sample may limit the predictors' performance for certain domains. Adding a small parallel sample to obtain 0k anchor point instances seems to be effective work around for this problem. 
\section{Conclusion}
In this work, we formulate a problem of Domain Adaptation Learning Curves prediction for NMT as instance-level learning framework. We demonstrate that it is possible to learn reasonable learning curve prediction model with a very small amount of NMT model instances via instance-level learning rather than corpus-level learning that most previous works rely on.  We propose a DaLC model relying on NMT encoder representations, combined with various instance and corpus-level features.  We show that such model is able to achieve good results with small amount of pretrained model instances. 
We perform in-depth analysis of the results for the domains where the predictor was less successful and conclude that the capacity of the predictor relying on the source-side sample only can be limited for some domains. Further analysis of characteristics of such domains could lead to better Domain Adaptation strategies. 
Finally, we believe it will be interesting to deepen the connections between Active Learning framework and Learning Curves prediction frameworks which could mutually help one another. 
\section{Acknowledgement}

The authors would like to thank Seunghyun S. Lim, Hyunjoong Kim and Stéphane Clinchant for valuable comments.
\bibliographystyle{acl_natbib}
\bibliography{anthology,custom}

\appendix
\newpage
\input{appendix}

\end{document}

%% file: appendix.tex
\clearpage
\section{NMT model training and domain adaptation hyper parameters}
\label{app:mt}
\subsection{Baseline}

 \paragraph{English-German} Our baseline models rely on \texttt{transformer-base} architecture \cite{vaswani2017attention}  trained on WMT-20 German-English dataset. For English-German direction we used inline casing \cite{berard2019} and tokenized text with BPE \cite{sennrich-etal-2016-neural} using 24K vocabulary size encoding joint vocabulary shared between two languages. We also share embedding parameters between source and target.  We use Fairseq toolkit \cite{ott2019fairseq} for training the baseline model with default training parameters for \texttt{transformer-base} architecture as proposed by ~\citet{vaswani2017attention}. Specifically, $M_{G}$ consists of six layers, 512 units and eight heads. We utilize the same Adam warm-up optimizer as in the original paper. Note that the baseline model for Finetuning and Adapter experiments in the same language direction are identical regardless of Domain Adaptation methods. 

\paragraph{Korean-English}
We follow the same experiment setting described in Appendix.~\ref{app:implem}. We utilize \texttt{transformer-big} architecture~\cite{vaswani2017attention} trained on the Ko-En direction. Training datasets of baseline model is constructed by using six categories datasets\footnote{We aggregate news, dialogue, colloquial, Korean culture, ordinance and website.} in AI-Hub~\footnote{\url{https://aihub.or.kr/aidata/87}}. The total number of training dataset is 1.51M parallel sentences. We set the size of BPE as 32K.

%

\subsection{Domain Adaptation}
Hyper-parameters used for Domain Adaptation models are reported in the Table \ref{tab:da_params}.
We use early stopping criteria on the validation loss performance in order to avoid overfitting to the data. Note that at this stage we used greedy decoding to compute chrF scores. We verified that the actual learning curves with greedy decoding behaves similarly to the learning curves with beam search. We stick to greedy decoding because we believe that beam search brings additional complexity to the performance prediction. We leave the impact of decoding method on the performance prediction for future work.

Table \ref{tab:da_bleu_chrf} and Table \ref{tab:da_bleu_chrf_ko_en} report the BLEU/mean chrF scores reached by each of our models on each language. We note that the model adapted via finetuning follows the same trend as the one adapted via adapter layers. Table \ref{tab:da_stats} reports best achieved loss for each domain and each anchor point, as well as the number of epochs required to reach it. One needs to keep in mind that the size of a single epoch varies across anchor points (it is 10 times bigger for 10k anchor point than for 1k anchor point). Note that loss is directly comparable between different domains and different anchor points since it corresponds to Cross-entropy loss relying on the same vocabulary.
\begin{table}[]
    \centering
    
    \begin{tabular}{c|c|c|c}
    \hline
                & En-De  & De-En &Ko-En\\\hline
        Learning rate & 0.001 & 0.001 &0.001 \\
         batch size & 4k & 2k  &2k\\
         \hline
    \end{tabular}
    \caption{Parameters used for Domain Adaptation of NMT system in each direction, where the batch size denotes the tokens per batch. En-De/FT and En-De/Adapter utilize the same learning rate and the batch size. }
    \label{tab:da_params}
\end{table}

\begin{table*}[]
    \centering
      \begin{tabular}{|c|cc|cc|cc|cc|cc|}
    \hline
         & \multicolumn{2}{c|}{IT} & \multicolumn{2}{c|}{Koran} & \multicolumn{2}{c|}{Law} & \multicolumn{2}{c|}{Medical} & \multicolumn{2}{c|}{Subtitles} \\
         & BLEU & chrF & BLEU & chrF & BLEU & chrF & BLEU & chrF & BLEU & chrF  \\\hline
         & \multicolumn{10}{c|}{German-English/FT} \\\hline
        Baseline& 29.2& 0.506& 10.2 & 0.335 & 27.8& 0.554 &32.7 & 0.585 & 20.9& 0.415\\\hline
        1K & 27.8& 0.498&13.3 &0.367 &34.6& 0.600 &36.6 & 0.629 & 21.2 & 0.393 \\
        10K &30.8 & 0.568& 20.2& 0.430& 40.1 & 0.641 & 45.0 & 0.702& 24.9& 0.439\\
        20K &33.5 & 0.596& 19.7 & 0.426& 42.6&  0.648& 47.7&  0.709&26.0 & 0.451\\
        100K &39.5 &0.658 & - & - & 49.0 & 0.696& 55.1&  0.753& 30.8& 0.500 \\
        
        \hline\hline
         & \multicolumn{10}{c|}{English-German/FT} \\\hline
        Baseline &32.1&0.544 & 11.9& 0.358 &36.8& 0.627 &34.5& 0.599 &21.8&0.472 \\\hline
        1K & 33.4&0.576 &16.3 & 0.393& 39.2 &0.643 &37.7& 0.634 &23.9&0.472\\
        10K & 34.6 & 0.605 &21.5 &0.441 & 42.3 &0.668 & 41.9&0.664 &25.2&0.482\\
        20K &35.9 &0.619 &23.4& 0.452&43.3& 0.675&42.6& 0.668 &25.2& 0.484\\
        100K & 38.2&0.653 & - & - & 46.7 &0.702  & 46.6&0.694 &25.6 & 0.490\\
        
        \hline
        \hline
         & \multicolumn{10}{c|}{German-English/Adapters} \\\hline
        Baseline & 29.2& 0.506& 10.2 & 0.335 & 27.8& 0.554 &32.7 & 0.585 & 20.9& 0.415 \\\hline
        1K & 30.0& 0.520 & 12.5& 0.341 &33.1 & 0.562& 33.0 & 0.613 &16.7 &0.374 \\
        10K & 30.2 & 0.562 &17.2& 0.387 & 38.6 &  0.614 & 41.9& 0.670 &21.5& 0.416\\
        20K & 31.3& 0.585& 19.0&0.401 & 40.0 & 0.629 & 44.2& 0.691& 23.5 & 0.445 \\
        100K & 35.2 &  0.638& - &- & 46.0& 0.673 & 51.1& 0.730& 25.4 &0.476 \\
        \hline
        
    \end{tabular}
    \caption{Domain Adaptation performance across domains for English-German and German-English models as measure either in BLEU or in chrF scores.}
    \label{tab:da_bleu_chrf}
\end{table*}

\begin{table*}[]
    \centering
      \begin{tabular}{|c|cc|cc|cc|cc|cc|}
    \hline
         & \multicolumn{2}{c|}{IT} & \multicolumn{2}{c|}{Koran} & \multicolumn{2}{c|}{Law} & \multicolumn{2}{c|}{Medical} & \multicolumn{2}{c|}{Subtitles} \\
         & Epoch & loss & Epoch & loss & Epoch & loss & epoch & loss & epoch & loss  \\
        
        \hline
         & \multicolumn{10}{c|}{English-German/FT} \\\hline
        1K & 4 & 3.718 & 12 & 4.228 & 4 &  3.033& 5 & 3.203 & 3 &  3.840\\
        10K & 12 & 3.502& 20& 3.680& 10 & 2.929 & 12 &  3.034 &9 &  3.784\\
        20K & 11 & 3.433 & 20 & 3.489 & 9 & 2.881 & 10 &  2.971 &7 &  3.762\\
        100K & 11 & 3.242 &- & -& 13 &  2.759 & 12 &  2.809 &6 &  3.691\\
        
        \hline
         
        
    \end{tabular}
    \caption{Domain Adaptation convergence statistics. We report the number of epochs when the Domain Adaptation reached the best validation loss, as well as the value of best validation loss. Note that each epoch corresponds to different amount of updates for different anchor point due to the difference in the in-domain samples used for DA. The loss however is comparable across different anchor points and different domains since it is always based on the same vocabulary.
}
    \label{tab:da_stats}
\end{table*}

\begin{figure*}
    \centering
    \includegraphics[width=\linewidth]{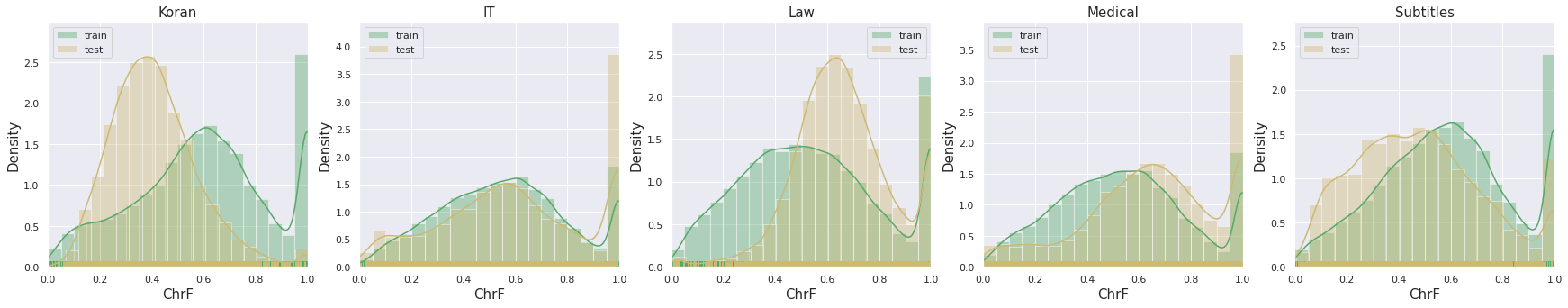}
    \caption{ChrF score distribution of training and test datasets across each experiment case in the English-German FT experiment. Each histogram shows the difference between the training and test ChrF score distributions in each least one out setting. For instance, the training distribution of Koran column obtains from all training anchor points of IT, Law, Medical and Subtitles domain. Test distribution of Koran column represents chrF score in all anchor points of the Koran domain.}
    \label{fig:dist_en_de}
\end{figure*}

\section{Implementation details of XGboost}
\label{app:xgboost}
We run the XG-boost regressor using XGBoost library~\footnote{https://github.com/dmlc/xgboost}. The learning rate of the XG-boost is set to 0.1. We follow the default regularization to alleviate the overfitting problem. The objective function of the XG-boost is RMSE. Regarding  the model parameters, the number of trees and the maximum depth of trees are 100 and 10, respectively.





\section{Computation time and memory usage}
\label{app:comp_cost}
\begin{table}[]
\resizebox{\columnwidth}{!}{\begin{tabular}{l|l|l|l}
Model            & Train time(s/epoch) & Test time(s/anchor) & params \\\hline
\texttt{exp3}             & 0.001               & 0.0002                         & 3      \\\hline
XGboost-corpus   & 0.23                & 0.0013                         &        100\\\hline
XGboost-instance & 1.44                & 0.0802                         &        1.03M\\\hline
DaLC             & 8.23                & 0.0914                         & 3.9M  \\\hline
\end{tabular}}
\caption{Computation time and memory usage of each model, where the test time is the time to inference one anchor point. }
\label{tab:computation_cost}
\end{table}
We compare the computation time and memory usage of DaLC with other models on the learning curve prediction task in Table~\ref{tab:computation_cost}. Although, corpus-level frameworks(i.e., XGboost-corpus and exp3) require less computation cost and memory resources, the performances of these models with small number of anchor points are very far from the real value described in Table~\ref{tab:full_results_ende}. The most computational heavy part of our model is the computation of anchor points used to train the predictor model. Anchor point implies training a Domain Adapted model with in-domain samples (200~300 sec in best case scenario of 1000 in-domain samples and up to 4h for larger in-domain samples). For this reason, instance-level frameworks are more practical in real-life scenarios as they achieve good performance with small amount of anchor points.  Note, that previous works \citet{xia-etal-2020-predicting, kolachina-etal-2012-prediction} relied on ~50-90 of MT model instances to train reliable corpus-level predictor. 

Considering the performance and total training time to obtain training anchor points, instance-level frameworks are feasible solutions to estimate the future performance without finetuning. In instance-level frameworks, DaLC is slightly slower than XGboost-instance but DaLC outperforms other approaches as shown in Table~\ref{tab:full_results_ende}. Moreover, DaLC is a very light in terms of parameter size compared to transformer base (65M). Our model can accurately forecast the finetuning performance with only 4\% of the total number parameters of \texttt{transformer base} without the need of computation resources for finetuning.  
\section{Performance on the Mixed domain }
\label{app:mix-ko-en}

\begin{table*}[]
    \centering
      \begin{tabular}{|c|cc|cc|cc|cc|cc|}
    \hline
         & \multicolumn{2}{c|}{Finace} & \multicolumn{2}{c|}{Social} & \multicolumn{2}{c|}{Sports} & \multicolumn{2}{c|}{Tech} & \multicolumn{2}{c|}{Travel} \\
         & BLEU & chrF & BLEU & chrF & BLEU & chrF & BLEU & chrF & BLEU & chrF  \\\hline
         & \multicolumn{10}{c|}{Korean-English/FT} \\\hline
        Baseline & 49.5 & 0.718 & 37.9 & 0.644 & 46.2 & 0.678 & 27.6 & 0.583 & 37.2 & 0.623 \\\hline
        1K       & 52.5 & 0.736 & 41.9 & 0.674 & 48.8 & 0.695 & 50.9 & 0.728 & 39.9 & 0.645 \\
        10K      & 54.6 & 0.749 & 44.5 & 0.691 & 51.2 & 0.712 & 56.9 & 0.768 & 41.4 & 0.655 \\
        20K      & 55.5 & 0.754 & 45.6 & 0.699 & 52.2 & 0.717 & 58.9 & 0.780 & 42.1 & 0.660 \\
        100K     & 58.0 & 0.768 & 48.5 & 0.718 & 54.7 & 0.732 & 63.2 & 0.807 & 43.8 & 0.672 \\
        \hline

    \end{tabular}
    \caption{Domain Adaptation performance across domains for Korean-English model as measure either in BLEU or in chrF scores.}
    \label{tab:da_bleu_chrf_ko_en}
\end{table*}

\begin{table}[]
\begin{tabular}{l|l|l|l}
\hline
                               & Mixed & TED  & Tech \\ \hline
\textit{exp3} & 0.231  & 0.297    & 0.136     \\ \hline
XGboost-Corpus                 & 0.170  & 0.070    & 0.229     \\ \hline
XGboost-Instance                 & 0.169  & 0.196    & 0.141     \\ \hline
DaLC                           & 0.105  & 0.112    & 0.099     \\ \hline

\end{tabular}
\caption{RMSE on the mixed domain experiments, where `Mixed' column means RMSE error on performance estimation in the mixed domain test sentences. `TED' and `Tech' column represents the performance on only test sentences on the corresponding domain in the mixing domain.}
\label{tab:mixed_domain}
\end{table}

\begin{table}[]
    \centering
      \begin{tabular}{c|c|c|c}
    \hline
         & Mixed & TED & Tech  \\\hline
         & \multicolumn{3}{c}{Korean-English/FT} \\\hline
        Baseline &0.503 & 0.423&0.583 \\\hline
        1K       & 0.562 & 0.427 &	0.697 \\
        10K      &0.590 & 0.431& 0.748\\
        20K      &0.596 & 0.429 & 0.763 \\
        100K     & 0.618 & 0.439 & 0.795 \\
        \hline

    \end{tabular}
    \caption{ChrF scores on each test set with a mixed Domain Adaptation case, where the models trained with uniformly sampled TED and Tech training dataset. Note that a Domain Adaptation model on each anchor point(i.e.,row-wise) is the same adaptions model. `Mixed' represents average chrF scores on test sentences of both TED and Tech domains.}
    \label{tab:da_chrf_mix_ko_en}
\end{table}


We evaluate our model under a loosely defined domain, such as mixing of  two domains. To simulate this scenario, we defines a mixed domain by aggregating Tech and TED domain. The samples $S_{n.d}$ of size $n$  in the mixed domain uniformly sampled from each domain. The test set of a mixed domain also is constructed with all test sentences of both domains.  We train DaLC with anchor points of Finance, Social, Sports and Travel (i.e., Except `Tech' domain) and then evaluate the performance on the mixed domain. Table~\ref{tab:da_chrf_mix_ko_en} summaizes the performance on each anchor point in the mixed domain experiment.

Table~\ref{tab:mixed_domain} demonstrates the RMSE errors across anchor points on the defined mixed domain and each domain. As previously, DaLC reaches lower (and more stable) prediction error compared to corpus-level models on averages. Thus, even the corpus-level can reach lower error in some cases (E.g., XGboost-corpus on TED) they still tend to be very close to predicting the mean score of the training data, and therefore their predictions are not stable across domains, while DaLC's predictions are more stable. 


\section{Predictor implementation details}
\label{app:implem}

The overall architecture of the predictor is described in Figure \ref{fig:arch}. Our prediction is composed of two parts, the encoder pooling component and the fusion layer component. The encoder pooling component is based on the multi-filter CNN architecture proposed by \citet{kim-2014-convolutional} which is widely used in text classification tasks. It encodes the sentence to latent feature $h_{i}$, which contains context information to predict performance. 

At first, the model obtains the encoder representation $e_{i}\in \mathcal{R}^{d}$ of the $i$-th token of the input $x={(x_{1},x_{2}... , x_{n})}$ from the encoder of the baseline model($M^G$), where $d$ and $n$ is the size of dimension and the length of $x$ . The size of hidden dimension $d$ is 512 in our implementation. Next, we apply the multiple convolution filters across the sequence in the same manner as the original paper, where the window sizes of convolution filters are 2,3 and 4. 
 We then obtain the output $h_{i} \in \mathcal{R}^{d}$ from the max pooling operation along the sequence of convolution outputs. To fuse all features, we concatenate the features of encoder representations $h_{i}$, the pre-computed domain-difficulty features $\phi_{DL}(x)$ and the corpus-level features $\xi(S_d)$. The fusion layer forecasts the performance based on the concatenated features, where the fusion layer is constructed with a five layer feed-forward neural network with non-linear activation functions. We utilize a non-linear function as ReLU, but the last layer of the feed-forward neural network utilize the Sigmoid function to change the output from zero to one. Learning is done via MSE loss between the predicted score and the ground-truth score.

\paragraph{Training hyperparameters}
To optimize the model, we use the learning rate as 0.001 with Adam optimizer. Moreover, the learning rate gradually decreases based on an exponential decay scheduler. All experiments stop the training based on the early stopping with patience as 10. Note that the parameters of $M^{G}$ encoder are fixed while training the predictor. 

\section{Overview of predictor training dataset creation}
\label{sec:overview_predictor_Train}

\begin{itemize}
    \item We start from the deduplicated train/dev/test splits.
    \item we sampled 1k, 10k, 20k or 100k samples from the train split
    \item we finetuned English-German baseline NMT system on each of those samples
    \item We then were able to compute instance-level chrF for each domain, at each anchor point for dev and test splits.
\end{itemize}

\section{Additional results}
\label{app:supp_results}
\begin{figure*}[ht!]
\centering
    \includegraphics[scale=0.33]{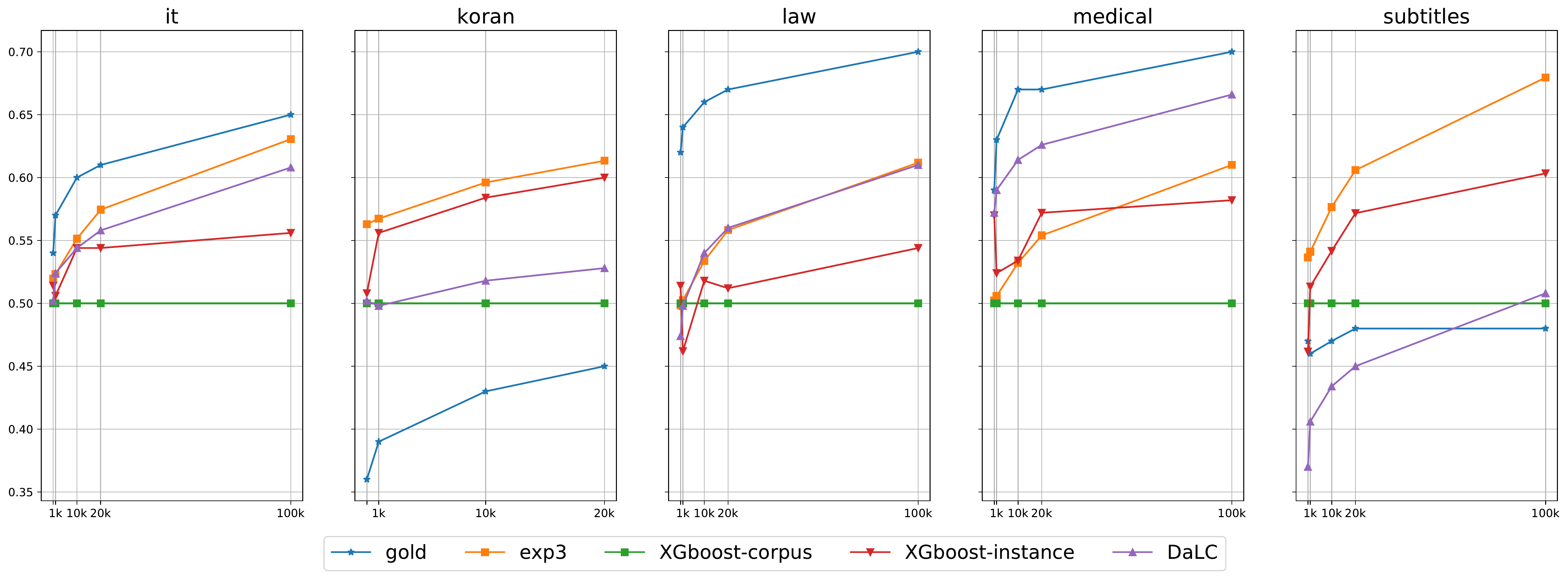}
    \caption{Learning curves provided by DaLC compared against \textit{gold} predictions,  baseline predictors ($exp_3$, XGboost-corpus, XGboost-instanc) for English-German.}
    \label{fig:lc_ende}
\end{figure*}

\begin{figure*}[!t]
    \centering
    \includegraphics[width=\linewidth]{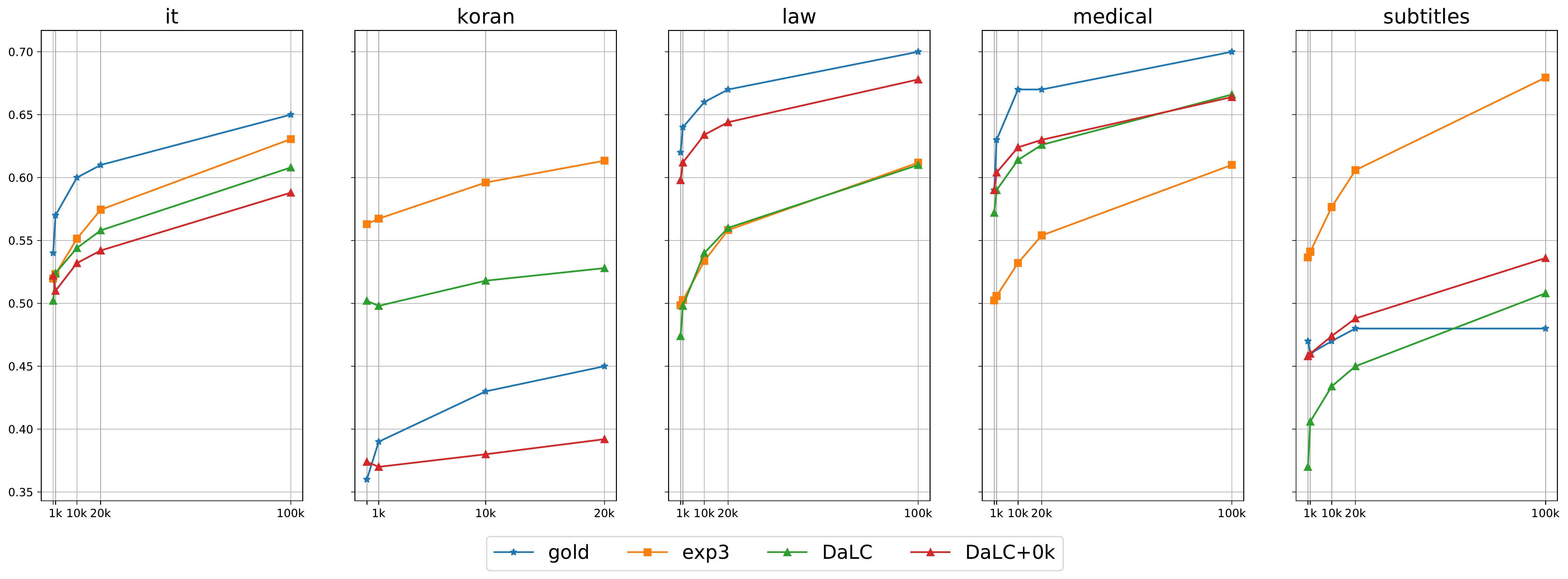}
    \caption{Comparison of learning curve prediction with DaLC and DaLC \texttt{+with 0k} on the English to German}
    \label{fig:dalc_0k_ende}
\end{figure*}    

\begin{figure*}[]
    \centering
    \includegraphics[angle=90,origin=c,scale=0.45]{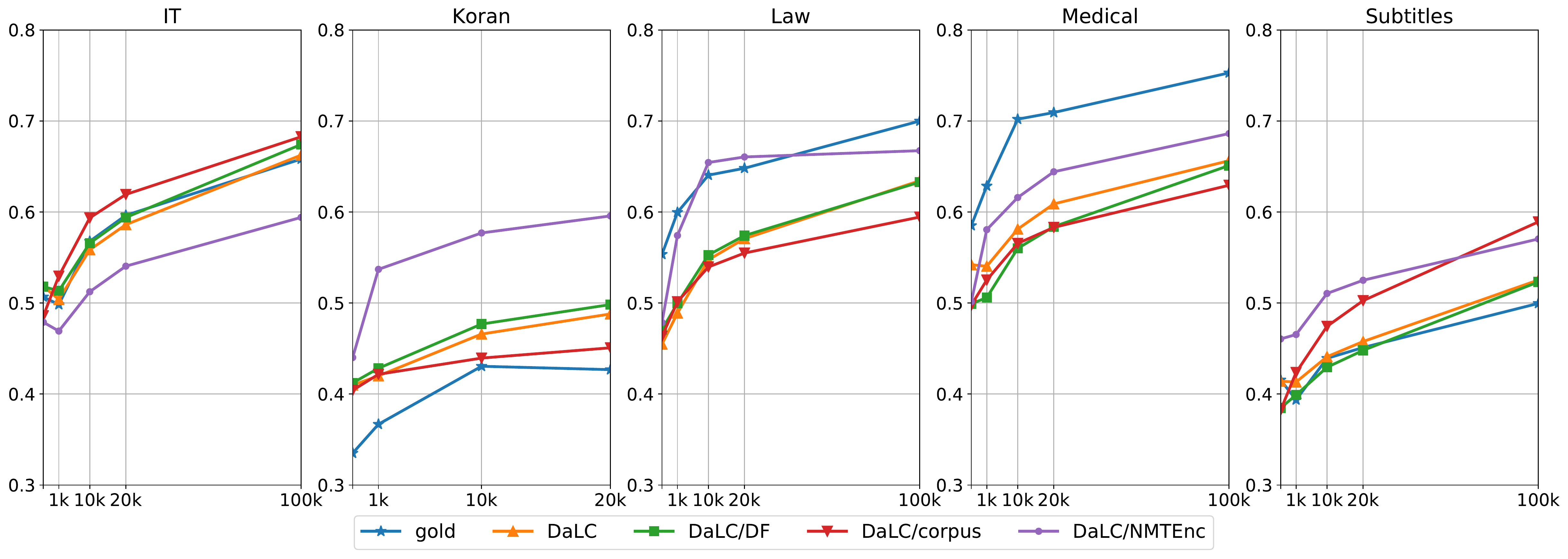}
    \caption{Learning curves for gold scores and predictors combining different components (please refer to section \ref{result_discussion} for details of those combinations.  Best viewed in color. }
    \label{fig:lc_full}
\end{figure*}

 Table \ref{tab:res_rmse_deen_ft} reports prediction errors (RMSE/MAE) across all available anchor points for all domains for  \textbf{DaLC}. Additionally, we provide results of the learning curves depending on the different combination components of DaLC in Figure \ref{fig:lc_full}.  Table \ref{tab:lc} reports the deviation from the gold predictions for each anchor point, each domain. Figure \ref{fig:lc_ende} is the learning curve prediction result in the English-German direction. We also report Table \ref{tab:with_0k_anchor} to demonstrate the detail RMSE errors with \texttt{+ with 0k} experiments described in Section \ref{sect:with_0k}

\section{Impact of the amount of domains}
\label{sec:extra_dom}

\citet{elsahar-galle-2019-annotate} have shown one can obtain better precision at performance drop prediction due to domain shift when the amount of training domains increases. Inspired by this finding, we extend the set of our training domains with additional domains (English-German language pair only): \textit{Bible}, \textit{Tatoeba}, \textit{Medline}, \textit{TAUS}, \textit{PatTR}, and \textit{MuchMore}. We provide some details of these various datasets in the Appendix \ref{app:extra_data}.

For each of these datasets, we created a predictor training dataset following the same settings as previously (whole procedure is outlined in the Appendix \ref{sec:overview_predictor_Train}).

We train DaLC predictor keeping the same test domains as previously, but extend corresponding training domains with newly introduced domains. 
We consider following extensions of the training domains:
(\textit{Bible, Medline, PatTR}), (\textit{TAUS, Tatoeba, MuchMore}) or All (6 new domains). The goal of this split is to decouple the differences in performance that are due to the amount of training data from those due to the nature of new domains added. Table \ref{tab:amount_domains} reports the results. We can see that additional domains indeed improve the overall error. We note that error decreases \textit{Law} domain, but error increases on \textit{Koran} domain. When looking at the result we see that the addition of new domains simply leads to higher predictions scores overall (which is due to overall high chrF scores in the additional domains). 

\section{Additional datasets}
\label{app:extra_data}
\begin{itemize}
    \item Bible, Tatoeba from Opus Website\footnote{\url{https://opus.nlpl.eu/}}
    \item Medline: consists of abstracts from scientific publications, distributed as part of WMT-20 Biomedical translation challenge\footnote{\url{http://www.statmt.org/wmt20/biomedical-translation-task.html}}
    \item TAUS\footnote{\url{https://md.taus.net/corona}}: TAUS Corona Crisis Corpora that consists of crawled documents related to Covid-19 crisis. 
    \item PatTR\footnote{\url{https://ufal.mff.cuni.cz/ufal_medical_corpus}}: patents related to medical domain
    \item MuchMore\footnote{\url{https://ufal.mff.cuni.cz/ufal_medical_corpus}}:  scientific medical abstracts obtained from the Springer Link web site.
\end{itemize}

\section{In-depth ablation study}
\label{sec:abblation}
Table \ref{tab:ablation_sub} reports the results for in-depth ablation study on Subtitles domain. Table \ref{tab:ablation_sub} shows the performance of the model after removing the corresponding feature from the input in the DaLC. The result shows that all features contribute to improve the performance in our task and empirically demonstrates the importance of each feature.  The normations correspond to
\begin{itemize}

\item MS : margin score,
\item LC: least-confidence score,
\item ATE  : average token entropy,
\item Labse: LaBSE based cosine similarity score,
\item c-length :  the average sentence length in characters,
\item n-token: the amount of tokens in $S_{d}^{s}$,
\item l-token : the average sentence length in tokens,
\item TTR : type token ratio in  $S_{d}^{s}$,
\item overlap:  vocabulary overlap ratio between $G$ and $S_{d}^{s}$, 
\item n-vocab :the number of unique tokens in $S_{d}^{s}$,
\end{itemize}
where the each term is mentioned at the Section \ref{sec:features}.

\begin{table*}[]
    \centering
    \begin{tabular}{|c|c|c|c|c|c|c|c|c|c|c|}
    \hline
        domain&\multicolumn{2}{c|}{Mean} & \multicolumn{2}{c|}{$exp_3$-fit} & \multicolumn{2}{c|}{XGboost-instance}&\multicolumn{2}{c|}{DaLC/DF} & \multicolumn{2}{c|}{DaLC} \\\hline
         & RMSE & MAE  & RMSE & MAE & RMSE & MAE &RMSE & MAE & RMSE & MAE  \\\hline
         koran& 0.328 & 0.164  & 0.292  &0.145 &0.201 & 0.200& 0.064&0.065 & \textbf{0.058}&\textbf{0.056} \\
         it& 0.073 & 0.028 &0.125 &0.049 &0.084 & 0.076&0.011 & 0.009&\textbf{0.009} & \textbf{0.009} \\
         medical& 0.366 & 0.162& 0.397 & 0.176 & 0.176& 0.172&0.117& 0.115& \textbf{0.094}&\textbf{0.082}  \\
         law& 0.248 & 0.108 & 0.305 &  0.134 & 0.085 & 0.079&0.058 & \textbf{0.047}& \textbf{0.057} &0.053 \\
         subtitles& 0.308 & 0.094    & 0.321 & 0.139& 0.126 & 0.120& 0.022& 0.019 & \textbf{0.015} & \textbf{0.007} \\
         \hline
    \end{tabular}
    \caption{Global Learning Curve prediction error: we report RMSE/MAE across all the anchor points available for each domain. De-En translation, Adaptation method: finetuning}
    \label{tab:res_rmse_deen_ft}
\end{table*}

\begin{table*}[]
   \centering
   \begin{tabular}{|c|c|c|c|c|c|}
         \hline
        &it & koran & law & medical & subtitles \\\hline
0& -0.011 (0.506) & -0.075 (0.335) & 0.082 (0.554) & 0.043 (0.585) & 0.001 (0.415)\\
1000& -0.005 (0.498) & -0.053 (0.367) & 0.072 (0.600) & 0.088 (0.629) & -0.020 (0.393)\\
10000& 0.010 (0.568) & -0.036 (0.430) & 0.039 (0.641) & 0.120 (0.702) & -0.002 (0.439) \\
20000& 0.002 (0.596) & -0.061 (0.426) & 0.029 (0.648) & 0.100 (0.709) & 0.007 (0.451) \\
100000& -0.004 (0.658) & - & 0.044 (0.696) & 0.096 (0.753) & -0.026 (0.500) \\
\hline

    \end{tabular}
   \caption{Learning Curves prediction results. En-De translation, Adaptation method: finetuning  Each cell reports gold \textit{mean chrF} score and prediction  model deviation from gold mean chrF ( prediction error ). }
    \label{tab:lc}
\end{table*}

\begin{table*}
    \centering
   \resizebox{\textwidth}{!}{\begin{tabular}{|c|ccccc|ccccc|}
    \hline
         & \multicolumn{5}{c|}{German-English/FT}  &\multicolumn{5}{c|}{English-German/FT}  \\\hline
          & IT& Koran & Law &  Medical& Subtitles  & IT & Koran & Law &  Medical & Subtitles \\\hline
         
        XGboost-instance&0.092 & 0.209 & 0.080   &0.176 & 0.151 &0.075& 0.158& 0.155& 0.105& 0.083\\
        DaLC & \textbf{0.006} & 0.051 & 0.050 & 0.102 & \textbf{0.017} & 0.064 & 0.111 & 0.128 & 0.032 & 0.050\\\hline
         & \multicolumn{10}{c|}{Adding 0k anchor point of the test domain } \\\hline
        XGboost + with 0k &0.900  &  0.144 & 0.073 & 0.156 &0.900 &  0.076& 0.152& 0.160 & 0.108& 0.086\\
        DaLC  + with 0k&0.007 & \textbf{0.032} & \textbf{0.029} & \textbf{0.060} & 0.024 & \textbf{0.059} & \textbf{0.028} & \textbf{0.023} & \textbf{0.016} & \textbf{0.035} \\
        \hline
    \end{tabular}}
    \caption{RMSE for DaLC predictor and XGboost baseline when adding 0k anchor point of a test domain as additional training data on English-German language pair. \texttt{+with 0k} represents the model utilizes the 0k anchor point of the given test domain in training. We report RMSE across all anchor points of the test domain excluding the 0k anchor point.  }
    \label{tab:with_0k_anchor}
\end{table*}

\begin{table*}[]

\resizebox{\textwidth}{!}{%
\begin{tabular}{|l|l|l|l|l|l|l|l|l|l|l|l|}
\hline
     & \multicolumn{1}{c|}{Full}    & \multicolumn{4}{c|}{$\phi_{DF}$}                                                                                         & \multicolumn{6}{c|}{$\xi$}                                                                                                                                                                          \\ \cline{2-12} 
     & \multicolumn{1}{c|}{DaLC} & \multicolumn{1}{c|}{wo MS} & \multicolumn{1}{c|}{wo LC} & \multicolumn{1}{c|}{wo ATE} & \multicolumn{1}{c|}{wo Labse} & \multicolumn{1}{c|}{wo c-length} & \multicolumn{1}{c|}{wo n-token} & \multicolumn{1}{c|}{wo l-token} & \multicolumn{1}{c|}{wo TTR} & \multicolumn{1}{c|}{wo overlap} & \multicolumn{1}{c|}{wo n-vocab} \\ \hline
RMSE & 0.0097                       & 0.0175                        & 0.0103                     & 0.0186                      & 0.0112                        & 0.0179                      & 0.0174                         & 0.0241                         & 0.0263                         & 0.0194                         & 0.0119                           \\
MAE  & 0.0085                       & 0.0141                        & 0.0121                     & 0.0149                      & 0.0095                        & 0.0153                      & 0.0147                         & 0.0176                         & 0.0176                         & 0.0159                         & 0.0102                           \\ \hline

\end{tabular}%
}
\caption{Ablation study of DaLC. AL with the subtitles domain, where the model train with the other domains case, such as koran, medical, it and law domains. `wo' indicates removal of corresponding feature from the input of the model. For example, `wo lf' indicates removing the least confidence score in the input.}
\label{tab:ablation_sub}
\end{table*}

\begin{table*}[]
    \centering
    \begin{tabular}{|c|c|c|c|c|c|c|}
    \hline
         & IT & Koran & Law & Medical & Subtitles & Avg \\\hline
         4 domains & 0.054& 0.114& 0.124& 0.032& 0.06& 0.077 \\
          + Bible + Medline + PatTR & 0.033& 0.128& 0.076& 0.031& 0.044& 0.062\\
          + TAUS + Tatoeba + MuchMore &  0.016& 0.124& 0.083& 0.012& 0.092& 0.065 \\
          + All  & 0.028& 0.136& 0.061& 0.018& 0.089& 0.066\\
          \hline
    \end{tabular}
    \caption{Impact of the number of domains on the predictors performance (En-De, FT)}
    \label{tab:amount_domains}
\end{table*}

